%% file: acl2023.tex
\pdfoutput=1

\documentclass[11pt]{article}

\usepackage[]{ACL2023}

\usepackage{times}
\usepackage{latexsym}
\usepackage{multirow}
\usepackage{url}
\usepackage{latexsym}
\usepackage{balance}
\usepackage{bm}
\usepackage{amssymb}
\usepackage{amsmath}
\usepackage{url}
\usepackage{makecell}
\usepackage{multirow}
\usepackage{subfig}
\usepackage{graphicx} 
\usepackage{color}
\usepackage{colortbl}
\usepackage{textcomp}
\usepackage{CJK}
\usepackage{enumitem}
\usepackage{booktabs}
\usepackage{microtype}
\usepackage{arydshln}
\usepackage{amsfonts}
\usepackage{eufrak}

\usepackage[T1]{fontenc}

\usepackage[utf8]{inputenc}

\usepackage{microtype}
\usepackage{pifont}
\usepackage{inconsolata}

\usepackage{booktabs}
\usepackage{tcolorbox}

\title{Self-Evolution Learning for Discriminative Language Model Pretraining}

\newcommand{\co}{$^{\dagger}$}
\author{%
  Qihuang~Zhong$^{1}$\thanks{~~~~Equal contribution.},
  Liang~Ding$^{2*}$,
  \textbf{Juhua~Liu}$^{3}$\thanks{~~Corresponding Authors: Juhua Liu (e-mail: liujuhua@whu.edu.cn), Bo Du (e-mail: dubo@whu.edu.cn)},
  \textbf{Bo~Du}$^{1}$\co,
  \textbf{Dacheng~Tao}$^{4}$ \\
  \fontsize{9.0pt}{\baselineskip}\selectfont $^{1}$ National Engineering Research Center for Multimedia Software, Institute of Artificial Intelligence, School of Computer Science \\ 
  \fontsize{9.0pt}{\baselineskip}\selectfont  and Hubei Key Laboratory of Multimedia and Network Communication Engineering, Wuhan University, China \\
  \fontsize{9.0pt}{\baselineskip}\selectfont $^{2}$ JD~Explore~Academy, China \quad $^{3}$ Research Center for Graphic Communication, Printing and Packaging, \\
  \fontsize{9.0pt}{\baselineskip}\selectfont and Institute of Artificial Intelligence, Wuhan University, China 
  \quad $^{4}$ University of Sydney, Australia \\
   \fontsize{9.0pt}{\baselineskip}\selectfont \texttt{\{zhongqihuang, liujuhua, dubo\}@whu.edu.cn},
   \texttt{\{liangding.liam, dacheng.tao\}@gmail.com}
}

\begin{document}
\maketitle
\input{sections/0_abstract.tex}
\input{sections/1_intro.tex}
\input{sections/2_related_work.tex}

\input{sections/3_methods.tex}
\input{sections/4_experiments.tex}
\input{sections/5_discussion.tex}
\input{sections/6_conclusion.tex} 

\bibliography{acl2023}
\bibliographystyle{acl_natbib}

\input{sections/7_appendix.tex}

\end{document}

%% file: sections/0_abstract.tex
\begin{abstract}

Masked language modeling, widely used in discriminative language model (\textit{e.g.}, BERT) pretraining, commonly adopts a random masking strategy. However, random masking does not consider the importance of the different words in the sentence meaning, where some of them are more worthy to be predicted. Therefore, various masking strategies (\textit{e.g.}, entity-level masking) are proposed, but most of them require expensive prior knowledge and generally train from scratch without reusing existing model weights. In this paper, we present \textit{Self-Evolution learning} ($\mathbb{SE}$), a simple and effective token masking and learning method to fully and wisely exploit the knowledge from data. $\mathbb{SE}$ focuses on learning the informative yet under-explored tokens and adaptively regularizes the training by introducing a novel \textit{Token-specific Label Smoothing} approach. Experiments on 10 tasks show that our $\mathbb{SE}$ brings consistent and significant improvements (+1.43$\sim$2.12 average scores) upon different PLMs. In-depth analyses demonstrate that $\mathbb{SE}$ improves linguistic knowledge learning and generalization.
\end{abstract}

%% file: sections/1_intro.tex
\section{Introduction}
\label{sec:intro}
Masked language modeling (MLM), which commonly adopts a random masking strategy to select the mask tokens, has become the \textit{de-facto} standard for discriminative pretrained language models (PLMs)~\cite{devlin2019bert,liu2019roberta,he2020deberta,joshi2020spanbert}. However, such a random masking process is usually criticized as being sub-optimal, as it allocates an equal masking rate for all tokens. In particular, the masked tokens are sometimes too easy to guess with only local cues or shallow patterns~\cite{joshi2020spanbert}, while the informative tokens that carry more critical linguistic knowledge may be neglected~\cite{church1990word,sadeq2022informask}.
For example, ``\textit{Bush}'' and ``\textit{Sharon}'' express more important meaning than ``\textit{a}'' in the sample sentence ``\textit{Bush held a talk with Sharon}''. MLM with predicting the above easy-to-guess tokens, \textit{e.g.}, ``a'', would lead to low data efficiency and sub-optimal model capability. 

To address this problem, various methods have been carefully designed to improve MLM via fully leveraging the training data~\cite{sun2019ernie,joshi2020spanbert,levine2020pmi}.
The common goal is to inject language prior knowledge into the pretraining process~\cite{cui2022lert,ding2021understanding}. Although empirically successful, there are still some limitations. 
First, they usually require annotation derived from off-the-shelf tools to select mask tokens, which is not only expensive but also too deterministic\footnote{The once-for-all prior is not suitable for different PLMs, \textit{e.g.}, under-explored word for BERT may already well-mastered by RoBERTa.}, and may cause error propagation from the third-party tool.
For instance, \citet{sun2019ernie} employ external linguistic tools, \textit{e.g.}, Stanford CoreNLP~\cite{manning2014stanford}, to annotate the entities.
Second, to ensure the effectiveness of the masking strategy, most previous works train PLM from scratch without reusing the existing models trained with vanilla MLM~\cite{sun2019ernie,joshi2020spanbert,levine2020pmi,sadeq2022informask}, which is wasteful and inefficient.

Thus, there \textbf{raises a question}:
\textit{whether we can strengthen the PLM capability and data efficiency through further learning from the informative yet under-explored tokens, where such tokens are determined by the existing PLM itself.
}
In fact, an off-the-shelf PLM already has the ability to determine the worthy and informative tokens that should be further exploited, as the representation of PLM generally can reveal good enough linguistic properties~\cite{hewitt-manning-2019-structural,swayamdipta-etal-2020-dataset}. For example, tokens that PLMs predict incorrect or low confidence are usually more hard-to-learn and challenging, which are essential for further training. Also, the conjecture to improve the off-the-shelf PLM is model-agnostic, green, and efficient, thus having the great potential to evolve any existing discriminative PLMs.

Motivated by this, we design a simple and effective \textit{Self-Evolution learning} ($\mathbb{SE}$) mechanism to improve the pretraining of discriminative PLMs. Specifically, the $\mathbb{SE}$ contains two stages: \ding{182}\textit{self-questioning} and \ding{183}\textit{self-evolution training}. In \textbf{stage~1}, the PLM is forced to locate the informative but under-explored tokens\footnote{We refer to those hard-to-learn tokens that are not learned well by PLMs as the informative but under-explored tokens.} from the pretraining data.
After locating these hard-to-learn tokens, we then encourage the PLM to learn from them in \textbf{stage~2}, where we basically follow the vanilla MLM to mask these tokens and then optimize the PLM by minimizing the loss between the predictions and one-hot labels.
It should be noted that due to the hard-to-learn properties, directly enforcing the PLM to fit the hard labels may lead to overfitting or overconfidence problem~\cite{miao-etal-2021-prevent}.
Inspired by the label smoothing (LS)~\cite{szegedy2016rethinking} that regularizes the learning by smoothing target labels with a pre-defined (static) prior distribution, we propose a novel \textit{Token-specific Label Smoothing} (TLS) approach. Our TLS considers both the precise hard label and, importantly, the \textit{easily-digestible}\footnote{Analogous to human learning behavior, it is often easier for humans to grasp new things described by their familiar knowledge~\cite{reder2016building}.} distribution that is adaptively generated by the PLM itself.

We validated our $\mathbb{SE}$ on several benchmarks including GLUE~\cite{wang2018glue}, SuperGLUE~\cite{wang2019superglue}, SQuAD2.0~\cite{rajpurkar2018know}, SWAG~\cite{zellers-etal-2018-swag} and LAMA~\cite{petroni2019language} over several PLMs: BRET~\cite{devlin2019bert}-\textsc{Base}, -\textsc{Large}, RoBERTa~\cite{liu2019roberta}-\textsc{Base}, and -\textsc{Large}. Experiments demonstrate the effectiveness and universality of our approach.
Extensive analyses confirm that $\mathbb{SE}$ effectively enhances the ability of PLMs on linguistic knowledge learning, model generalization and robustness.
\paragraph{Contributions}Our main contributions are:
\begin{itemize}
    \item We propose $\mathbb{SE}$ to strengthen the MLM-based PLMs, where our mechanism does not require external tools and enjoys a simple recipe: continue pretraining with $\mathbb{SE}$.
    \item We design a novel token-specific label smoothing approach for regularization, which adopts the token-specific knowledge-intensive distributions to adaptively smooth the target labels.
    \item Extensive experiments show that our $\mathbb{SE}$ could significantly and robustly evolve a series of backbone PLMs, up to +2.36 average score improvement on GLUE benchmark upon RoBERTa.
\end{itemize}


%% file: sections/2_related_work.tex
\begin{figure*}[ht]
    \centering
    \includegraphics[width=0.95\textwidth]{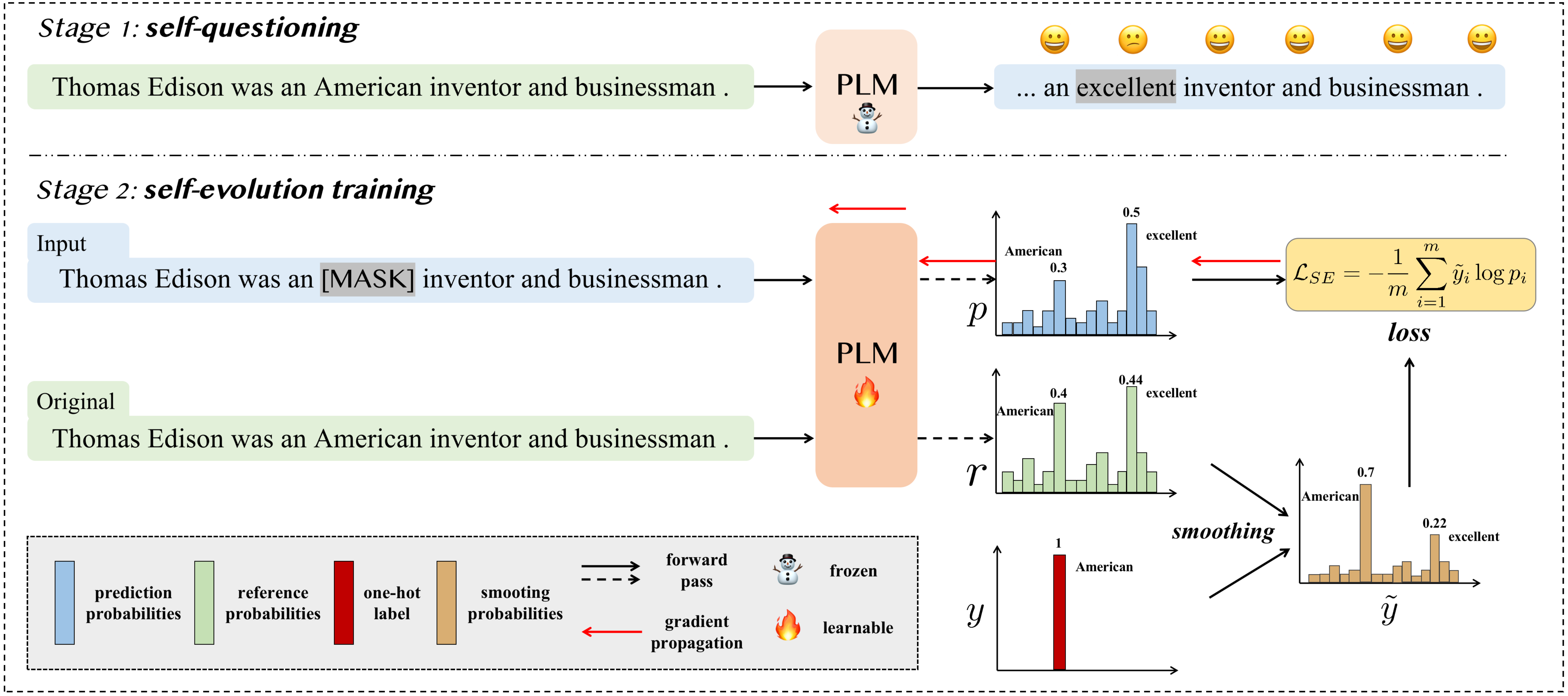}
    \caption{\label{fig:self-evolution}Overview of the proposed $\mathbb{SE}$ mechanism, which contains two stages: \ding{182} using an existing PLM to locate the informative yet under-explored tokens and \ding{183} encouraging the PLM to robustly learn from these tokens via a token-specific label smoothing approach. 
    }
\end{figure*}

\section{Related Works}
\label{sec:related}
In recent years, we have witnessed numerous discriminative PLMs~\cite{devlin2019bert,liu2019roberta,he2020deberta,sun2019ernie,joshi2020spanbert} that achieved tremendous success in various natural language understanding (NLU) tasks. Although the discriminative PLMs vary in terms of pretraining data or model architecture, they are commonly based on MLM loss function. MLM mechanism is pioneered in BERT~\cite{devlin2019bert} that uses a random masking strategy to mask some tokens, and then enforces the PLM to learn to recover word information from the masked tokens. Obviously, the vanilla MLM is a linguistic-agnostic task, as the random masking procedure does not integrate linguistic knowledge explicitly, which is sub-optimal. Thus, several previous studies attempt to improve MLM by exploring a diverse of linguistically-motivated masking strategies, such as entity-level masking~\cite{sun2019ernie}, span-level masking~\cite{joshi2020spanbert}, N-grams masking~\cite{levine2020pmi}, \textit{etc.}, to fully leverage the pretraining data. 

Although achieving remarkable performance, these strategies still have some limitations. First, their implementations are relatively complex, as they usually require annotation derived from external models or tools to select tokens for masking.
Even for the unsupervised PMI-masking~\cite{sadeq2022informask}, it is still expensive to measure the pointwise mutual information for pretrain-level large-scale data, and the annotated labels are static, while our $\mathbb{SE}$ could obtain dynamic annotations via given existing PLMs.
Second, in order to ensure the effectiveness of masking strategy, most previous works~\cite{sun2019ernie,joshi2020spanbert,levine2020pmi,sadeq2022informask} train the language models from scratch without reusing the existing PLMs trained with vanilla MLM, which is wasteful and inefficient.

Along the same research line, in this paper, we improve the MLM-based PLMs with a novel self-evolution learning mechanism. Instead of training a PLM from scratch based on a carefully-designed and complex masking strategy, our mechanism aims to strengthen the PLM's capability and data efficiency by further learning from the informative yet under-explored tokens, which are determined by the existing PLM itself.

%% file: sections/3_methods.tex
\section{Methodology}
\label{sec:method}
\subsection{Preliminary}
Given a sentence $S=\{t_1,t_2,...,t_n\}$ with $n$ tokens, MLM first randomly selects some percentage of the input tokens and replaces them with a special mask symbol \texttt{[MASK]}. Suppose that there are $m$ masked tokens and $\{k_1,k_2,...,k_m\}$ is the set of masked positions, we can denote the masked tokens as $M=\{t_{k_1},t_{k_2},...,t_{k_m}\}$. Let $S'$ denote the masked sentence, we can feed $S'$ into the model and obtain the last hidden layer representations as $H \in \mathbb{R}^{n \times d}$ ($d$ is the hidden size), and a subset of representations w.r.t masked positions as $H^m \in \mathbb{R}^{m \times d}$. Subsequently, the input word embedding matrix $E \in \mathbb{R}^{V \times d}$ ($V$ is the vocabulary size) is used to project the hidden representations into vocabulary space. Lastly, we can get the normalized prediction probabilities for each masked token as:
\begin{align}
    p_i=\text{softmax}(H^m_i E^T+ b),
\end{align}
where $p_i \in \mathbb{R}^V$ and $ i\in\{1,2,...,m\}$.
Finally, given the one-hot labels $y_i$, we use the cross-entropy loss to optimize the MLM task:
\begin{align}
    \mathcal{L}_{MLM}=-\frac{1}{m}\sum_{i=1}^m y_i \log p_i
    \label{mlm_loss}
\end{align}

\subsection{Self-Evolution Learning for PLMs}
\label{sec:se_method}
In this part, we introduce our $\mathbb{SE}$ mechanism in detail. At its core, $\mathbb{SE}$ is to enforce the existing PLM to further learn from the informative yet under-explored tokens, which are wisely determined by the PLM itself.
Figure~\ref{fig:self-evolution} illustrates the process of $\mathbb{SE}$ mechanism, which contains two stages: (1) \textit{self-questioning} and (2) \textit{self-evolution training}. 

\paragraph{\ding{182} Self-questioning Stage.}
The goal of this stage is to select the informative yet under-explored tokens, \textit{i.e.}, these hard-to-learn tokens that the PLMs do not learn well during the previous pretraining. However,
\textit{how to select these target tokens?} Inspired by the finding of the representations of the off-the-shelf PLM on individual tokens can reveal good enough linguistic properties~\cite{hewitt-manning-2019-structural,swayamdipta-etal-2020-dataset}, we hereby propose to straightforwardly leverage the behavior of PLMs to wisely select target tokens in this stage.
Specifically, we mainly focus on two important properties, \textit{i.e.}, \textbf{\textit{correctness}} (accuracy) and \textbf{\textit{confidence}} (the probability output that the model assigns to the prediction), as the tokens that PLMs predict incorrect or low confidence are usually more hard-to-learn and worthy for further exploring~\cite{guo2017calibration,park2022calibration}.
Based on the above two properties, we introduce two simple metrics to estimate the learning value of tokens:

\textbf{Correctness-based metric.} In practice, we first feed the original sentence $S$ into the existing frozen PLM and enforce it to output the prediction probabilities $p_i$ ($i\in\{1,2,...,n\}$) for each token. Given the one-hot labels $y_i$ ($i\in\{1,2,...,n\}$), we calculate the cross-entropy loss (\textit{i.e.}, correctness) for each token position (denoted as $\{l_1,l_2,...,l_n\}$). Then, we set a loss threshold $\mathcal{T}_l$ and select the tokens that exceed $\mathcal{T}_l$ as the target tokens, \textit{i.e.}, $M=\{t_i|l_i>\mathcal{T}_l\}$ where $i\in\{1,2,...,n\}$.

\textbf{Confidence-based metric.} Similarly, we can measure the confidence of tokens and use it as the metric. Different from the above process, in this metric, we compute the entropy of $p_i$ as the confidence for each token (denoted as $\{e_1,e_2,...,e_n\}$). Intuitively, the tokens with high entropy value are hard-to-learn, as the PLM predict them with low confidence towards the gold labels. Also, an entropy threshold $\mathcal{T}_e$ is used to select the target tokens, \textit{i.e.}, $M=\{t_i|e_i>\mathcal{T}_e\}$\footnote{
In practice, 
$\mathcal{T}_l$ and $\mathcal{T}_e$ are empirically set as 0.1 and 1, respectively. The detailed analyses are shown in Appendix~\ref{appendix_parameter_analysis}. 
}.

\paragraph{\ding{183} Self-evolution Training Stage.} After estimating these hard-to-learn tokens, we can then choose them for masking and encourage the PLM to learn from them. Intuitively, we can follow the vanilla MLM process to optimize the PLM by minimizing the loss between the predictions and one-hot labels, as implemented in Eq.~\ref{mlm_loss}. However, due to the hard-to-learn properties of these tokens, directly enforcing the PLM to fit the hard labels may lead to overfitting or overconfidence problem~\cite{miao-etal-2021-prevent,li2022pre}. To tackle this issue, in this stage, inspired by the label smoothing (LS) regularization approach~\cite{szegedy2016rethinking}, we further propose a novel \textit{token-specific label smoothing} (TLS) approach to adaptively regularize the training and improve the generalization of PLMs.

Mathematically, in LS approach, it minimizes the cross-entropy between modified label distribution $y'_i$ and the model output $p_i$, where $y'_i$ is the smoothed label distribution formulated as:
\begin{align}
y'_i=(1-\lambda)*y_i+\lambda*u_i,
\label{eq_ls}
\end{align}
where $u_i$ is a fixed distribution that is usually a uniform distribution, and $\lambda$ is a weighting factor. Furthermore, following~\citet{yuan2020revisiting}, we reformulate the loss function of LS as:
\begin{align}
\mathcal{L}_{LS}=(1-\lambda)*H(y, p)+\lambda*D_{kl}(u,p),
\end{align}
where $H$ denotes the ordinary cross-entropy loss and $D_{kl}$ denotes the KL divergence loss. We can regard $D_{kl}(u,p)$ as a knowledge distillation process, where $u$ corresponds to a virtual teacher to guide the student model (\textit{i.e.}, the PLM). Obviously, it is sub-optimal as $u$ hardly provides enough linguistic information to guide the training of PLM. 

Motivated by this, in our TLS, we design a more informative prior distribution to smooth the labels. 
Specifically, inspired by human learning behavior (it is often easier for humans to grasp new things described by their familiar knowledge~\cite{reder2016building}), we improve the $D_{kl}$ supervision with a more easily-digestible and informative distribution that is adaptively generated by the PLM itself. In other words, $D_{kl}$ can be recast as a self-distillation process, where the virtual teacher distribution is acquired from the student model itself. In practice, for each masked position $k_i$, in addition to the prediction probabilities $p_i$ on the corrupted $S'$, we also feed the original sentence $S$ into the current PLM and regard the corresponding probabilities as the reference probabilities $r_i$\footnote{It is noteworthy that, although the $r_i$ may not be much close to the ground-truths, the linguistic information contained in $r_i$ is potentially beneficial for further learning.}. 
Then, similar to Eq.~\ref{eq_ls}, we can obtain the smoothed label $\Tilde{y}_i$ via:
\begin{align}
\Tilde{y}_i=(1-\lambda)*y_i+\lambda*r_i
\label{label_smoothing}
\end{align}
Lastly, we use the cross-entropy as the loss function in the $\mathbb{SE}$ training stage, as follows:
\begin{align}
    \mathcal{L}_{SE}=-\frac{1}{m}\sum_{i=1}^m \Tilde{y}_i \log p_i
    \label{se_loss}
\end{align}

%% file: sections/4_experiments.tex
\input{tables/main_results.tex}

\section{Experiments}
\label{sec:experiments}
\subsection{Tasks and Datasets}
We follow many previous studies~\cite{zhong2022panda,zhong2022toward,zhong2023revisiting,zhong2023bag} and conduct extensive experiments on various NLU tasks, including a diversity of tasks from GLUE~\cite{wang2018glue} and SuperGLUE~\cite{wang2019superglue} benchmarks, \textit{i.e.}, linguistic acceptability (CoLA), natural language inference (RTE, CB), paraphrase (MRPC), question answering (BoolQ), word sense disambiguation (WiC) and causal reasoning (COPA). Additionally, we also evaluate on three knowledge-intense tasks, which require the ability of commonsense knowledge reasoning, \textit{i.e.}, SQuAD2.0~\cite{rajpurkar2018know}, SWAG~\cite{zellers-etal-2018-swag} and LAMA~\cite{petroni2019language}.
In practice, we report the performance with Accuracy (``\textit{Acc.}'') metric for most tasks, except the Matthew correlation (``\textit{Mcc.}'') for CoLA, the F1 and Exact Match (``\textit{EM}'') scores for SQuAD2.0, and the Mean Reciprocal Rank (``\textit{MRR}'') scores for LAMA.
We report the averaged results over 10 random seeds to avoid stochasticity.
The details of all tasks and datasets are provided in Appendix~\ref{appendix_data}.

\subsection{Implementation Details}
\paragraph{Pre-training.} We employ the representative BRET~\cite{devlin2019bert}-\textsc{Base}, -\textsc{Large}, RoBERTa~\cite{liu2019roberta}-\textsc{Base}, and -\textsc{Large} as the backbone discriminative PLMs, and implement our methods in a continued pretraining manner. For pretraining settings, we follow the original papers~\cite{devlin2019bert,liu2019roberta} and use the same pretraining corpus and (most of) hyper-parameters\footnote{Notably, for the continued pretraining process, we use 1/10 of the learning rate in the original paper as the initial one.} (e.g., batch size and the maximum length of the input sentence), respectively.

Especially, as suggested by~\citet{liu2019roberta}, we do not use the next sentence prediction (NSP) objective during BERT pretraining. For our methods, we continue pretraining the backbone PLMs with 2 epochs. Additionally, for reference, we train the PLMs with the vanilla MLM for the same steps and refer to them as the baselines.

\paragraph{Fine-tuning.} The learning rate is selected in \{1e-5, 2e-5, 3e-5, 5e-5\}, while the batch size is in \{12, 16, 32\} depending on tasks. The maximum length of the input sentence is 384 for SQuAD2.0 and 256/512 for other tasks.  We use AdamW~\cite{loshchilov2018decoupled} as the optimizer, and set the $\beta_2$ and weight decay as 0.98 and 0.01, respectively. All experiments are conducted on NVIDIA A100 GPUs.
The detailed hyper-parameters are provided in Appendix~\ref{appendix_parameters}.

\paragraph{Compared Methods.} For references, we compare our $\mathbb{SE}$ method with other cutting-edge counterparts. Specifically, taking the BERT$\rm_{\texttt{base}}$ as the baseline, we use the following masking strategies to further improve its performance:
\begin{itemize}
    \item Entity-level masking: following~\citet{sun2019ernie}, we mask the named entities in the sentence and enforce the model to predict them.
    \item Span-level masking: as done in~\cite{joshi2020spanbert}, we randomly select spans from the sentence based on a geometric distribution and mask the selected span.
    \item PMI-based masking: similar to~\cite{sadeq2022informask}, we use PMI to identify a set of contiguous (informative) N-grams and mask them.
    \item Self-questioning masking\footnote{The main difference from our full $\mathbb{SE}$ is that it does not involve the self-evolution training process of \textbf{stage 2}.}: We adopt our \textbf{stage 1} to select the hard-to-learn tokens and directly follow the vanilla MLM to mask them and predict the one-hot labels.
\end{itemize}
Notably, for a fair comparison, we implement all these methods in a continual pretraining manner, same to the settings of our $\mathbb{SE}$.

\input{tables/main_results2.tex}
\input{tables/lama.tex}

\input{tables/ablation_metric.tex}

\subsection{Main Results}
\paragraph{$\mathbb{SE}$ surpasses the previous carefully-designed masking strategies.} Results on GLUE and SuperGLUE benchmarks are shown in Table~\ref{tab:main1}. Compared with the baseline BERT$\rm_{\texttt{base}}$, all masking strategies bring the average performance gains, proving the necessity of improving MLM. Among all these methods, our proposed self-questioning masking achieves the relatively better performance on many tasks, confirming the effectiveness of using the PLMs themselves to select the hard-to-learn tokens. More encouragingly, with the help of self-evolution training, our final BERT-SE$\rm_{\texttt{base}}$ can achieve further performance improvements. These results can prove the superiority of our $\mathbb{SE}$.

\paragraph{$\mathbb{SE}$ brings consistent and significant performance improvements among all PLMs.} 
In addition to the results upon BERT$\rm_{\texttt{base}}$, we also apply our method on more discriminative PLMs and report the results in Table~\ref{tab:main1}.
Compared with the baselines, $\mathbb{SE}$ brings consistent and significant performance improvements across all BERT/RoBERTa model sizes. Specifically, for Base and Large RoBERTa models, $\mathbb{SE}$ brings 2.36\% and 2.03\% relative gains in overall score respectively. Also, the gain for BERT is up to 2.24\%. These results prove the effectiveness and universality of our $\mathbb{SE}$.


\paragraph{$\mathbb{SE}$ enhances the ability of knowledge learning.} For the knowledge-intense tasks, \textit{i.e.}, SQuAD2.0 and SWAG, we report the results in Table~\ref{tab:main2}. With the help of $\mathbb{SE}$, all PLMs consistently achieve better performance. Specifically, the performance improvements on SQuAD2.0 in terms of EM and F1 are up to 0.71\% and 0.64\%, respectively. 
Besides QA tasks that require to be fine-tuned, we conduct experiments on a widely-used factual knowledge probing task, \textit{i.e.}, LAMA~\cite{petroni2019language}, 
to verify whether $\mathbb{SE}$ improves the ability of PLMs on commonsense knowledge.
We report the results in Table~\ref{tab_lama}. Based on the powerful RoBERTa, $\mathbb{SE}$ still brings significant improvements, \textit{i.e.} +3.8 average score, to the knowledge-learning ability of PLMs.

\begin{figure}[t]
    \centering
    \includegraphics[width=0.4\textwidth]{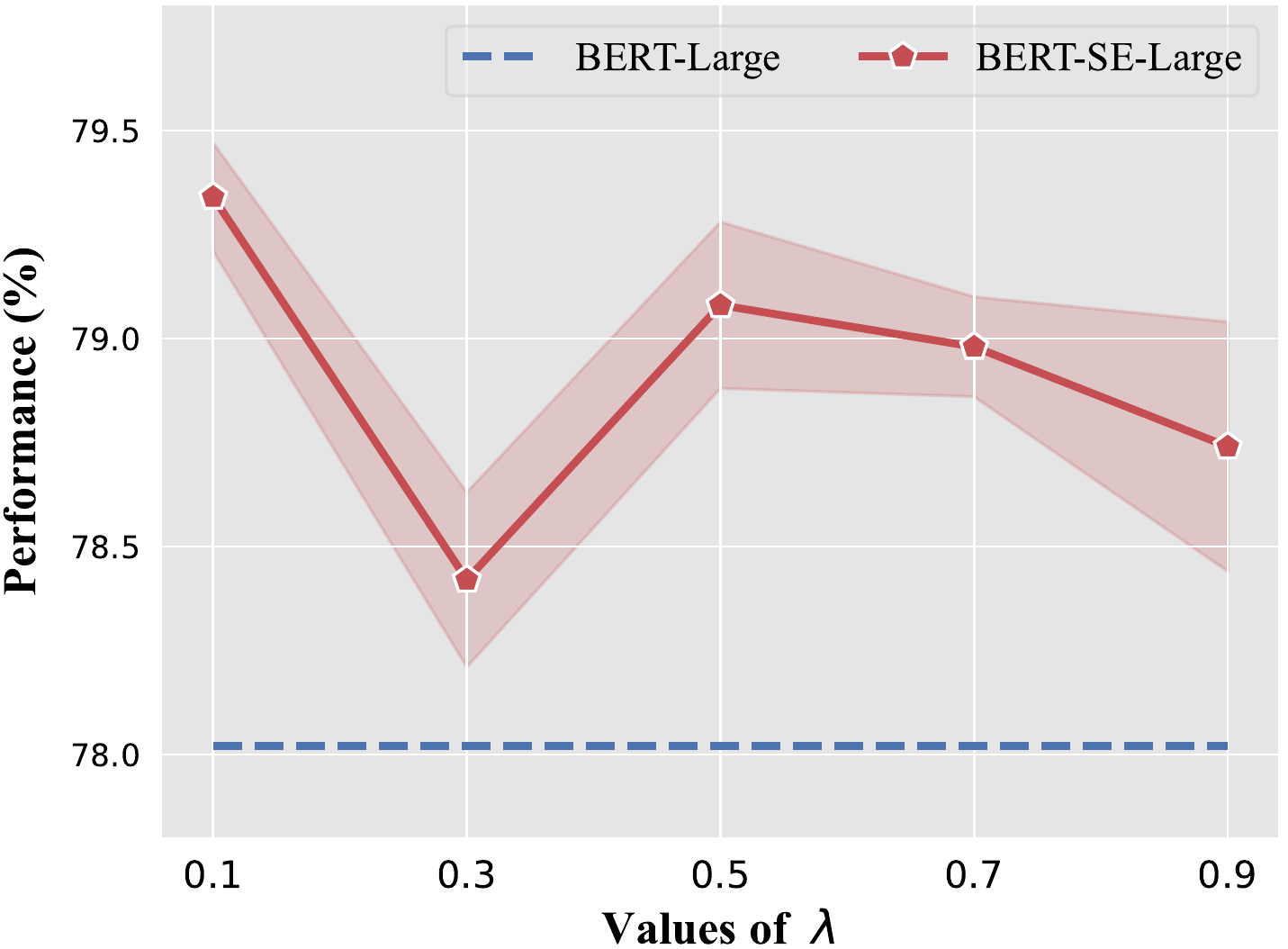}
    \caption{Parameter analysis of $\lambda$ on BERT-SE$\rm_{\texttt{large}}$.}
    \label{fig:ablation2}
\end{figure}

\subsection{Ablation Study}
We evaluate the impact of each component of our $\mathbb{SE}$, including \textit{i}) token-selecting metrics, \textit{ii}) token-specific label smoothing approach, \textit{iii}) coefficient $\lambda$, and \textit{iv}) more $\mathbb{SE}$ iterations.

\paragraph{Impact of Token Selecting Metrics.}
\label{sec_abalation_metric}
As mentioned in \S\ref{sec:se_method}, we introduce several metrics to select the hard-to-learn tokens in the self-questioning stage. Here, we conduct experiments to analyze the impact of different metrics. Specifically, for reference, we compare the ``Correctness-based'' and ``Confidence-based'' metrics\footnote{Our preliminary study shows the non-complementarity between two token-selecting metrics, we compute their vocabulary distribution difference and give evidence at Appendix~\ref{appendix_statistic_analysis}.}
with a simple alternative, \textit{i.e.}, ``randomly selecting''. Results in Table~\ref{tab_ablation_metric} show that 1) although the ``randomly selecting'' performs worst, it still outperforms the continually trained baseline, showing the effectiveness of the self-evolution training. 2) both our proposed metrics ``Correctness-based'' and ``Confidence-based'' achieve significantly better performance, confirming our claim that learning on informative yet under-explored tokens can strengthen the capability of PLMs and data efficiency. Notably, the correctness-based metric outperforms the confidence-based metric in most cases, thus leaving as our default setting in $\mathbb{SE}$.

\input{tables/ablation_tls.tex}
\paragraph{Impact of Token-specific Label Smoothing.} A key technology in our $\mathbb{SE}$ is the TLS, which uses the token-specific smoothed label to adaptively guide training. To verify its effectiveness, we conduct experiments and present the results in Table~\ref{tab_ablation_tls}. We show that 1) the vanilla label smoothing approach equipped $\mathbb{SE}$ could easily outperform the continuously trained backbone, showing the superiority of our $\mathbb{SE}$ framework, and importantly, 2) our TLS could further improve the results by a large margin against vanilla LS equipped $\mathbb{SE}$, \textit{e.g.} averaging +0.71, indicating the effectiveness of TLS.

\paragraph{Impact of Coefficient $\lambda$.} The factor
$\lambda$ in Eq.~\ref{label_smoothing}, which is used to control the ratio of label smoothing, is an important hyper-parameters. In this study, we analyze its influence by evaluating the performance with different $\lambda$ spanning \{0.1, 0.3, 0.5, 0.7, 0.9\} on several GLUE tasks. Figure~\ref{fig:ablation2} illustrates the average results. Compared with the baseline, our $\mathbb{SE}$ consistently brings improvements across all ratios of $\lambda$, basically indicating that the performance of $\mathbb{SE}$ is not sensitive to $\lambda$. More specifically, the case of $\lambda = 0.1$ performs best, and we thereby use this setting in our experiments.


\input{tables/iterativeSE.tex}

\begin{figure}[t]
    \centering
    \includegraphics[width=0.4\textwidth]{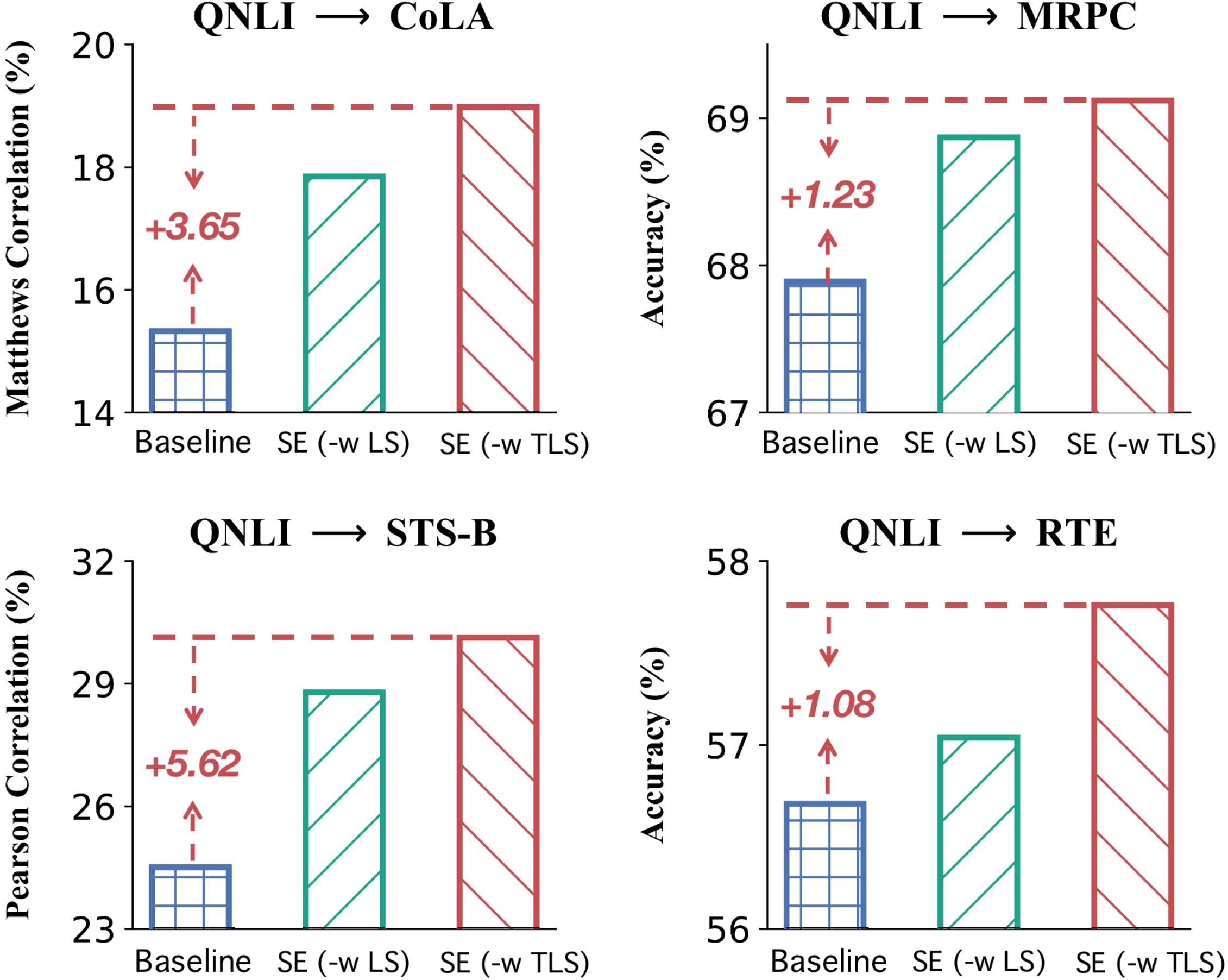}
    \caption{\label{fig:task_tranfer}Analysis of task generalization. The model is fine-tuned on the QNLI task and transferred to four different tasks. We can see that $\mathbb{SE}$ consistently brings better generalization compared with its  counterparts.}
    \label{fig:task_transfer}
\end{figure}

\begin{figure*}[ht]
	\centering
	\includegraphics[width=0.89\textwidth]{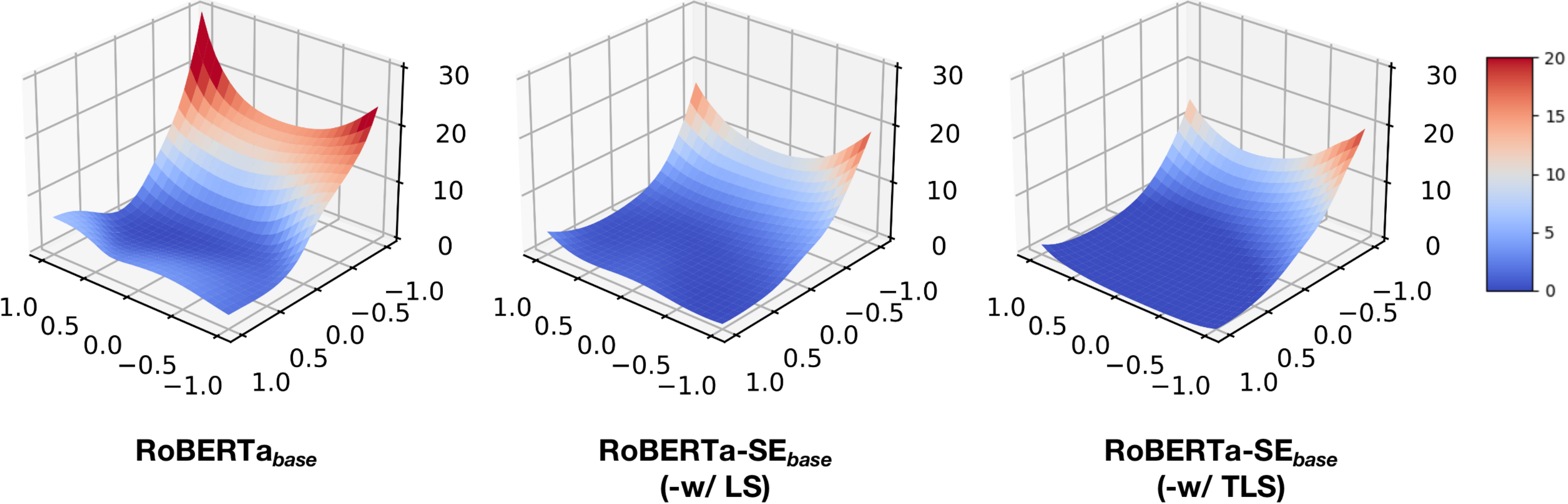} 
	\caption{The 3D loss surface comparison between baseline, $\mathbb{SE}$ (``-w/ vanilla LS'')  and $\mathbb{SE}$ (``-w/ TLS'') methods applied to RoBERTa$\rm_{\texttt{base}}$. Note that the PLMs are fine-tuned on the CoLA task. 
	}
	\label{fig:3d_loss}
\end{figure*}

\begin{figure}[ht]
    \centering
    \includegraphics[width=0.45\textwidth]{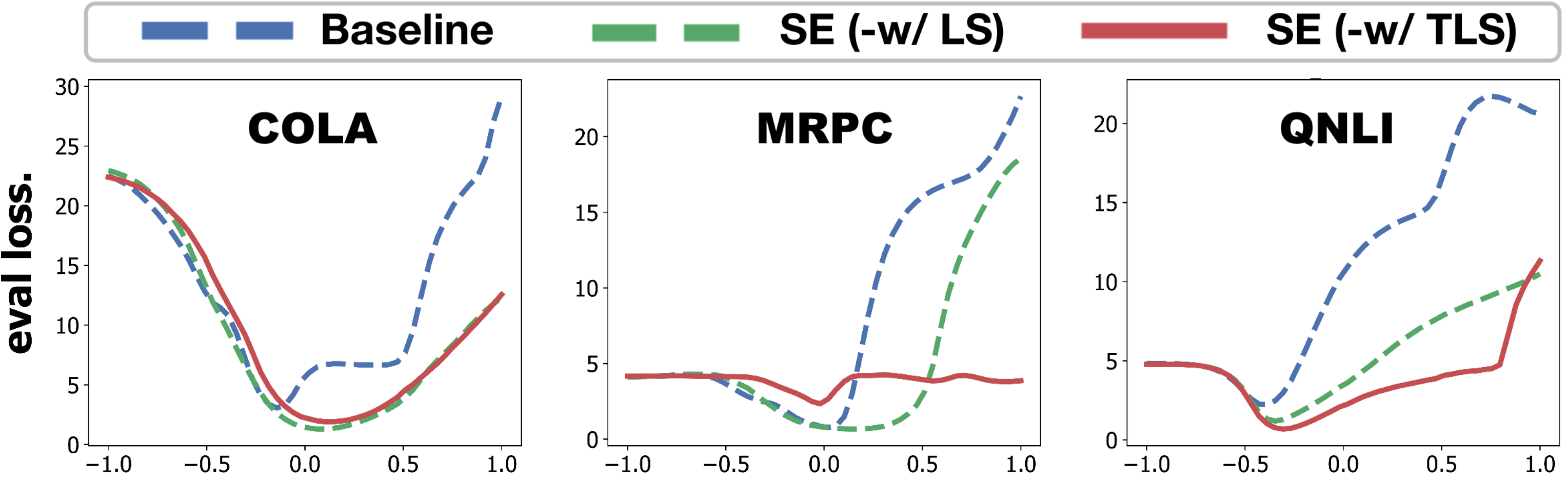}
    \caption{1D visualization of loss landscapes of RoBERTa$\rm_{\texttt{base}}$ models fine-tuned on different tasks. }
    \label{fig:1d_loss}
\end{figure}

\paragraph{Impact of More $\mathbb{SE}$ Iterations.}
Researchers may doubt whether $\mathbb{SE}$ can be further augmented by performing the self-questioning and token-specific label smoothing with already evolved PLMs that own better representations. That is, whether more iterations (denoted as ``$N$'') further enhance $\mathbb{SE}$? 
To answer this question, we continuously train the PLMs with more $\mathbb{SE}$ iterations and report the performance of several GLUE tasks in Table~\ref{tab:iterations}. As seen, increasing the iterations improves the performance but the gain margin is insignificant. Given that increasing $N$ costs more, we suggest using $\mathbb{SE}$ for only one iteration to achieve a better trade-off between costs and performance.

%% file: tables/main_results.tex
\begin{table*}[]
\centering
\scalebox{0.91}{
    \begin{tabular}{lccccccccc}
    \toprule
    \multicolumn{1}{c}{}                         &\bf CoLA                        &\bf MRPC                                              &\bf RTE                         &\bf BoolQ         &\bf CB            &\bf WiC           &\bf COPA          & \multicolumn{2}{c}{\bf Score}      \\
    \cmidrule(lr){2-4} \cmidrule(lr){5-8} \cmidrule(lr){9-10} 
    \multicolumn{1}{l}{\multirow{-2}{*}{\bf Method}} & \textit{Mcc.}               & \textit{Acc.}                             & \textit{Acc.}               & \textit{Acc.} & \textit{Acc.} & \textit{Acc.} & \textit{Acc.} & \textit{\underline{Avg.}} & \textit{$\Delta$ ($\uparrow$)} \\
    \midrule
    \multicolumn{10}{l}{\textit{Performance of Different Masking Strategies}}  \\ \midrule
    BERT$\rm_{\texttt{base}}$                               & 62.33                        & 88.97                                                & 76.89                        & \textbf{75.05}          & 85.71          & 66.77          & 63.00          & \underline{74.10}          & --              \\
    \hdashline
    \quad -w/ Entity-level masking & 60.06 & 88.73 & 76.53 & 74.77 & 87.50 & 66.61 & 65.00 & 74.17          & \textcolor[RGB]{0,176,80}{+0.07}             \\
    \quad -w/ Span-level masking & 61.41 & 88.48 & \textbf{78.34} & 74.28 & 87.50 & 67.40 & 65.00 & 74.63         & \textcolor[RGB]{0,176,80}{+0.53}             \\
    \quad -w/ PMI-based masking & 61.09 & 88.24 & 76.90 & 74.25 & 87.50 & 66.61 & 65.00 & 74.23           & \textcolor[RGB]{0,176,80}{+0.13}             \\
    \quad -w/ Self-questioning & \textbf{63.78} & 87.99 & \textbf{78.34} & 74.13 & 85.71 & \textbf{67.87} & \textbf{66.00} & 74.83         & \textcolor[RGB]{0,176,80}{+0.73}             \\
    \textbf{BERT-SE}$\rm_{\texttt{base}}$                         & 63.63 & \textbf{89.50} & 77.98 & 74.37 & \textbf{89.29} & 67.40 & \textbf{66.00} & \textbf{\underline{75.45}} & \textcolor[RGB]{0,176,80}{\textbf{+1.35}}            \\
    \midrule
    \multicolumn{10}{l}{\textit{Performance upon More Discriminative PLMs}}  \\ \midrule
    BERT$\rm_{\texttt{large}}$                               & 63.00                        & 87.25                                              & 83.80                        & 78.40          & 91.07         & 67.24          & 72.00          & \underline{77.54}          & --              \\
    \textbf{BERT-SE}$\rm_{\texttt{large}}$                           & \textbf{65.66}               & \textbf{88.23}                                              & \textbf{85.20}               & \textbf{80.18}          & \textbf{92.86}          & \textbf{68.34}          & \textbf{78.00}          & \textbf{\underline{79.78}}          & \textcolor[RGB]{0,176,80}{\textbf{+2.24}}            \\
    \hdashline
    RoBERTa$\rm_{\texttt{base}}$                            & 62.00                        & \textbf{90.20}                                                & 83.12                        & 78.72          & 83.93          & 69.12          & 70.00          & \underline{76.72}          & --              \\
    \textbf{RoBERTa-SE}$\rm_{\texttt{base}}$                         & \textbf{62.11}                        & 89.71                                               & \textbf{84.12}               & \textbf{79.39}          & \textbf{92.86}          & \textbf{71.40}          & \textbf{74.00}          & \textbf{\underline{79.08}}          & \textcolor[RGB]{0,176,80}{\textbf{+2.36}}            \\
    \hdashline
    RoBERTa$\rm_{\texttt{large}}$                            & 64.73                        & 90.69                                               & 88.44                        & 84.37          & 91.07          & 69.90          & 78.00          & \underline{81.03}          & --              \\
    \textbf{RoBERTa-SE}$\rm_{\texttt{large}}$                       & \textbf{67.80}               & \textbf{91.91}                                    & \textbf{90.25}                        & \textbf{84.56}          & \textbf{96.40}          & \textbf{70.53}          & \textbf{80.00}          & \textbf{\underline{83.06}}          & \textcolor[RGB]{0,176,80}{\textbf{+2.03}}            \\       
\bottomrule    
\end{tabular}
}
\caption{Comparison between our $\mathbb{SE}$ and the vanilla method applied to all PLMs on the combination of GLUE and SuperGLUE benchmarks. Average scores on all tasks are \underline{underlined}. The best results are given in \textbf{bold}. ``$\Delta$'' denotes the improvement of $\mathbb{SE}$ methods compared to the baseline PLMs. 
}
\label{tab:main1}
\end{table*}

%% file: tables/main_results2.tex
\begin{table}[t]
\scalebox{0.87}{
    \begin{tabular}{lcccc}
\toprule
    \multicolumn{1}{l}{\multirow{2}{*}{\bf Method}} & \multicolumn{2}{c}{\bf SQuAD2.0}    &\bf  SWAG           &\bf \multirow{2}{*}{\underline{Avg.}} \\
    \cmidrule(){2-4}
    \multicolumn{1}{c}{}                        & \textit{EM}             & \textit{F1}             & \textit{Acc.}           &                       \\
    \midrule
    BERT$\rm_{\texttt{base}}$                              & 72.18          & 75.07          & 77.53          & \underline{74.93}                 \\
    \textbf{BERT-SE}$\rm_{\texttt{base}}$                           & \textbf{72.89}          & \textbf{75.64} & \textbf{77.91}          & \textbf{\underline{75.48}}        \\
    \hdashline
    BERT$\rm_{\texttt{large}}$                            & 81.35          & 84.38          & 83.40          & \underline{83.04}                 \\
    \textbf{BERT-SE}$\rm_{\texttt{large}}$                       & \textbf{81.94} & \textbf{85.00} & \textbf{83.61}          & \textbf{\underline{83.52}}       
\\ \midrule
    RoBERTa$\rm_{\texttt{base}}$                              & 78.79          & 81.92          & 79.69          & \underline{80.13}                 \\
    \textbf{RoBERTa-SE}$\rm_{\texttt{base}}$                           & \textbf{79.41}          & \textbf{82.55} & \textbf{79.88}          & \textbf{\underline{80.61}}        \\
    \hdashline
    RoBERTa$\rm_{\texttt{large}}$                            & 84.70          & 87.65          & 84.34          & \underline{85.56}                 \\
    \textbf{RoBERTa-SE}$\rm_{\texttt{large}}$                       & \textbf{85.03} & \textbf{87.93} & \textbf{84.54}          & \textbf{\underline{85.83}} 
    \\

\bottomrule    
\end{tabular}
}
\caption{Performance on SQuAD2.0~\cite{rajpurkar2018know} and SWAG~\cite{zellers-etal-2018-swag} dev sets. 
}
\label{tab:main2}
\end{table}

%% file: tables/lama.tex
\begin{table}[t]
\centering
\small
\scalebox{0.84}{
\begin{tabular}{lcccc}
\toprule
\multicolumn{1}{l}{\multirow{2}{*}{\textbf{Method}}} & \multicolumn{3}{c}{\textbf{Google-RE (LAMA)}}                                                     & \multirow{2}{*}{\textbf{\underline{Avg.}}} \\ \cmidrule(lr){2-4}
\multicolumn{1}{c}{}                        & date-birth      & place-birth     & place-death     &                       \\ \midrule
RoBERTa$\rm_{\texttt{base}}$                               & 5.51                 & 11.52                & 2.68                 & \underline{6.57}                \\
\bf RoBERTa-SE$\rm_{\texttt{base}}$                           & \textbf{6.35}        & \textbf{15.16}       & \textbf{9.61}                       & \textbf{\underline{10.37}}        \\ 
\bottomrule
\end{tabular}
}
\caption{Performance of our $\mathbb{SE}$ on LAMA~\cite{petroni2019language} to probe the factual knowledge.
}
\label{tab_lama}
\end{table}

%% file: tables/ablation_metric.tex
\begin{table*}[]
\centering
\scalebox{0.9}{
\begin{tabular}{cccccc}
\toprule
\textbf{Method} & \textbf{BERT}$\rm_{\texttt{base}}$             & \textbf{BERT}$\rm_{\texttt{large}}$              & \textbf{RoBERTa}$\rm_{\texttt{base}}$            & \textbf{RoBERTa}$\rm_{\texttt{large}}$   & \textbf{\underline{Avg.}}     \\
\midrule
Baseline                    & 74.10                    & 77.54                    & 76.73                    & 81.03    & \underline{77.35}                 \\ 
\hdashline
\multicolumn{6}{l}{Selecting metrics in \textit{self-questioning} stage}  \\
\multicolumn{1}{l}{\quad -w/ randomly selecting}  &73.85 (\textcolor{red}{-0.25})   &78.28 (\textcolor[RGB]{0,176,80}{+0.74})   &77.09 (\textcolor[RGB]{0,176,80}{+0.36})   &81.64 (\textcolor[RGB]{0,176,80}{+0.61})  &\underline{77.72 (\textcolor[RGB]{0,176,80}{+0.37})}  \\	
\multicolumn{1}{l}{\quad -w/ Correctness-based}                                 & 75.45 (\textcolor[RGB]{0,176,80}{+1.35})                              & \textbf{79.78 (\textcolor[RGB]{0,176,80}{+2.24)}}                                          & \textbf{79.08 (\textcolor[RGB]{0,176,80}{+2.35)}}                                   & \textbf{83.06 (\textcolor[RGB]{0,176,80}{+2.03)}}                   & \underline{\textbf{79.34 (\textcolor[RGB]{0,176,80}{+1.99)}}}                \\
\multicolumn{1}{l}{\quad -w/ Confidence-based}                      & \textbf{75.77 (\textcolor[RGB]{0,176,80}{+1.67)}}                    & 78.88 (\textcolor[RGB]{0,176,80}{+1.34})                                & 77.86 (\textcolor[RGB]{0,176,80}{+1.13})                                & 82.46 (\textcolor[RGB]{0,176,80}{+1.43})                      & \underline{78.74 (\textcolor[RGB]{0,176,80}{+1.39)}}    \\
\bottomrule
\end{tabular}
}
\caption{Ablation study of different metrics used to select the hard-to-learn tokens in $\mathbb{SE}$, evaluated on the combination of GLUE and SuperGLUE benchmarks. For simplicity, we show the overall score here. The full results and analyses about the superiority of the correctness-based metric can be found in Appendix (Table~\ref{tab:all_results}\&\ref{tab:statistic}).}
\label{tab_ablation_metric}
\end{table*}

%% file: tables/ablation_tls.tex

\begin{table}[t]
\centering
\scalebox{0.8}{
\begin{tabular}{lcc}
\toprule
\multirow{2}{*}{\bf Method} &\bf GLUE/SGLUE &\bf SQuAD/SWAG \\ \cmidrule{2-3}
\multicolumn{1}{c}{} & \textit{Avg. ($\Delta \uparrow$)} & \textit{Avg. ($\Delta \uparrow$)} \\ \midrule
RoBERTa$\rm_{\texttt{base}}$ & 76.73 & 80.13 \\ \hdashline
\multicolumn{3}{l}{RoBERTa-SE$\rm_{\texttt{base}}$} \\
\quad -w/ vanilla LS & 78.37 (\textcolor[RGB]{0,176,80}{+1.64}) & 80.37 (\textcolor[RGB]{0,176,80}{+0.24}) \\
\quad -w/ TLS (Ours) & \textbf{79.08 (\textcolor[RGB]{0,176,80}{+2.35})} & \textbf{80.61 (\textcolor[RGB]{0,176,80}{+0.48})} \\
\bottomrule
\end{tabular}
}
\caption{Ablation study of our TLS approach. ``-w/ vanilla LS'' and ``-w/ TLS (Ours)'' refer to using the vanilla and our proposed token-specific label smoothing approaches in $\mathbb{SE}$ mechanism, respectively. Full results are shown in Appendix (Table~\ref{tab:all_results2}).}
\label{tab_ablation_tls}
\end{table}

%% file: tables/iterativeSE.tex
\begin{table}[]
    \centering
    \small
    {
    \begin{tabular}{lccc}
    \toprule
    {\bf GLUE}                  & {\bf $N=1$} & {\bf $N=2$} & {\bf $N=3$}\\
    \midrule
    CoLA	&  63.63   &63.59 &63.60 \\
    MRPC	&  89.50   & 88.23 &88.97 \\
    RTE     &  77.98   &79.42 &78.70 \\ \hdashline
\textit{Avg.} ($\Delta \uparrow$) &  \textcolor[RGB]{0,176,80}{+0.97}   &\textcolor[RGB]{0,176,80}{+1.02} &\textcolor[RGB]{0,176,80}{+1.03} \\ 
    \bottomrule
    \end{tabular}}
    \caption{Performance for different iterations $N$ on BERT-SE$_{\texttt{base}}$. ``\textit{Avg.} ($\Delta \uparrow$)'' indicates the relative improvement against the vanilla BERT$_{\texttt{base}}$.}
    \label{tab:iterations}
\end{table}

%% file: sections/5_discussion.tex
\section{Discussion}
\label{sec:discussion}
To better understand $\mathbb{SE}$, we conduct extensive analyses to discuss whether it gains better generalization/ robustness and knowledge-learning ability.

\subsection{Does $\mathbb{SE}$ Bring Better Generalization?}
We examine from two perspectives: \textit{i}) measuring the cross-task zero-shot performance, and \textit{ii}) visualizing the loss landscapes of PLMs.

\paragraph{Task Generalization.} 
The performance of out-of-domain (OOD) data is widely used to verify the model generalization~\cite{Wang2022UnderstandingAI,ding2022redistributing}. Thus, we follow~\citet{xu2021raise,zhong2022improving} and evaluate the performance of PLMs on several OOD data. In practice, we first fine-tune RoBERTa$\rm_{\texttt{base}}$ models trained with different methods (including ``Baseline'', ``$\mathbb{SE}$ (-w/ LS)'', and ``$\mathbb{SE}$ (-w/ TLS)'') on the QNLI task, and then inference on other tasks, \textit{i.e.}, CoLA, MRPC, STS-B, and RTE. The results are illustrated in Figure~\ref{fig:task_transfer}.
We observe that ``$\mathbb{SE}$ (-w/ TLS)'' consistently outperforms the other counterparts. To be more specific, compared with baseline, our $\mathbb{SE}$ brings a +2.90 average improvement score on these tasks, indicating that \textbf{\textit{our $\mathbb{SE}$ boosts the performance of PLMs on OOD data.}}

\paragraph{Visualization of Landscape.} To have a close look, we visualize the loss landscapes of different RoBERTa$\rm_{\texttt{base}}$ models fine-tuned on the CoLA task. In practice, we first show the 3D loss surface results in Figure~\ref{fig:3d_loss} following the ``filter normalized'' setting in~\cite{li2018visualizing,zan2022complementarity}. 
As seen, $\mathbb{SE}$-equipped PLMs show flatter smoother surfaces compared with the vanilla.
To closely compare the differences of ``$\mathbb{SE}$ (-w/ LS)'' and ``$\mathbb{SE}$ (-w/ TLS)'' in the loss landscape, we follow~\citet{he2021effectiveness} to plot the 1D loss curve on more tasks
in Figure~\ref{fig:1d_loss}. We find that through detailed 1D visualization, our optimal setting ``$\mathbb{SE}$ (-w/ TLS)'' shows a flatter and optimal property.
\textbf{\textit{These results prove that $\mathbb{SE}$ can smooth the loss landscape and improve the generalization of PLMs effectively.}}

\begin{figure*}[ht]
	\centering
	\includegraphics[width=0.95\textwidth]{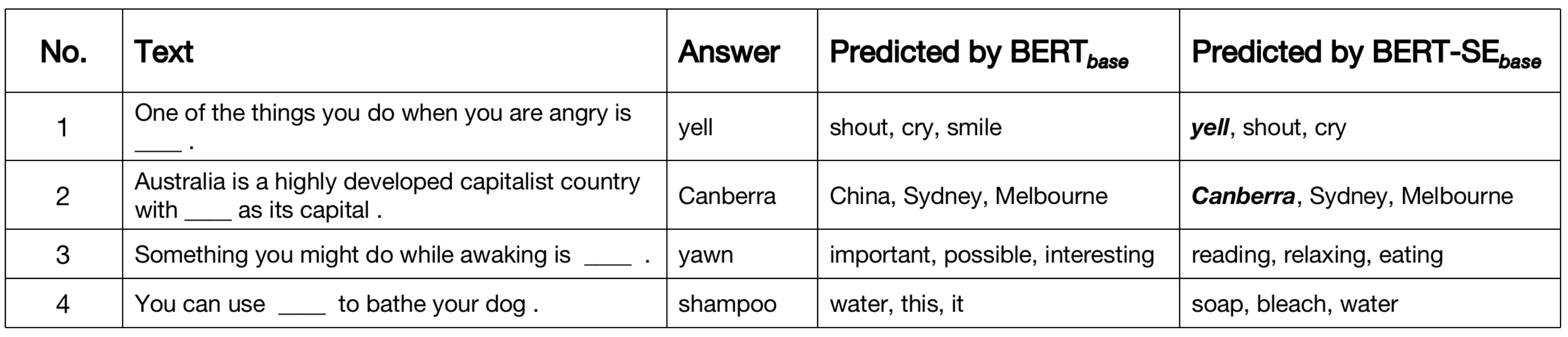} 
	\caption{Cloze test comparison between BERT$\rm_{\texttt{base}}$ and  BERT-SE$\rm_{\texttt{base}}$. The correct predictions are in \textbf{bold}.
	}
	\label{fig:cloze}
\end{figure*}

\subsection{Cloze Test}
To verify whether $\mathbb{SE}$ enforces the PLMs to learn from the informative tokens, we follow~\citet{sun2019ernie} and apply the Cloze test~\cite{taylor1953cloze} to evaluate the knowledge learning ability of PLMs. For each test sample, we first remove the informative token and then enforce the PLMs to infer what it is. Some cases are shown in Figure~\ref{fig:cloze}.

In case 1 and case 2, both BERT$\rm_{\texttt{base}}$ and BERT-SE$\rm_{\texttt{base}}$ can successfully predict the type of masked tokens according to the contexts. However, with the help of the $\mathbb{SE}$ mechanism, BERT-SE$\rm_{\texttt{base}}$ performs more correctly on filling in the slot. Dramatically, in case 3, the baseline BERT$\rm_{\texttt{base}}$ makes unreasonable predictions. One possible reason is that the baseline PLM only learns the shallow pattern and fails to understand the meaning of the context. Additionally, due to the unsatisfactory ability of the baseline PLM on commonsense reasoning, the baseline PLM also predicts strangely in case 4. Different from the baseline, while BERT-SE$\rm_{\texttt{base}}$ does not predict the completely correct tokens in case 3 and case 4, it can capture deep patterns and make more reasonable predictions. \textbf{\textit{In general, these cases prove that $\mathbb{SE}$ indeed improves the knowledge-learning ability of PLMs.}}

\paragraph{\large{\ding{43}} More analyses in Appendix} In addition to the above discussions, we conduct more related analyses and show them in Appendix, \textit{e.g.}, parameter analyses on $\mathcal{T}_l$ and $\mathcal{T}_e$ (Appendix~\ref{appendix_parameter_analysis}), robustness analysis based on the empirical results on AdvGLUE~\cite{wang2021adversarial} (Appendix~\ref{appendix_advglue}), and non-complementarity analysis between token-selecting metrics (Appendix~\ref{appendix_statistic_analysis}). Please refer to Appendix for more details.

%% file: sections/6_conclusion.tex
\section{Conclusion}
\label{sec:conclusion}
In this paper, we propose a simple and effective self-evolution ($\mathbb{SE}$) learning mechanism to improve the existing discriminative PLMs by fully exploiting the knowledge from data. $\mathbb{SE}$ follows two stages, \textit{i.e.}, \textit{self-questioning} and \textit{self-evolution training}, and can be used to evolve any MLM-based PLMs with a simple recipe: continue pretraining with $\mathbb{SE}$.
We empirically demonstrated the effectiveness and universality of the $\mathbb{SE}$ on a series of widely-used benchmarks. Further analyses show our approach improves the generalization, robustness, and knowledge-learning ability.
We hope our work could facilitate more research on how to improve existing trained models after all the previous PLM weights are expensive and knowledgeable. 


\section*{Limitations}
Our work has several potential limitations.
First, given the limited computational budget, we only validate our self-evolution learning on the Large and Base sizes. It will make our work more convincing if scaling the experiments up to the larger model size and training corpus. 
On the other hand, besides the improved commonsense knowledge learning ability, we believe that there are still other abilities, \textit{e.g.}, mathematical word problems, of PLMs that can be improved by our method, which are not fully explored in this work.

\section*{Ethics and Reproducibility Statements}
\paragraph{Ethics} We take ethical considerations very seriously, and strictly adhere to the ACL Ethics Policy. This paper focuses on higher data and model efficiency for discriminative pretrained language models, but not capturing the privacy knowledge. Both the pretraining datasets and models used in this paper are publicly available and have been widely adopted by researchers. Therefore, we believe that this research will not pose ethical issues.

\paragraph{Reproducibility} We will publicly release our code in \url{https://github.com/WHU-ZQH/SE4PLMs} to help reproduce the experimental results of this paper. 

\section*{Acknowledgements}
We are grateful to the anonymous reviewers and the area chair for their insightful comments and suggestions.
This work was supported
in part by the National Natural Science Foundation of China under Grants 62225113 and 62076186, and in part by the Science and Technology Major Project of Hubei Province (Next-Generation AI Technologies) under Grant 2019AEA170. The numerical calculations in this paper have been done on the supercomputing system in the Supercomputing Center of Wuhan University. 

%% file: sections/7_appendix.tex
\appendix
\section{Appendix}
\label{sec:appendix}
\input{tables/dataset.tex}

\input{tables/advglue.tex}

\subsection{Details of Tasks and Datasets}
\label{appendix_data}
Here, we introduce the descriptions of all downstream tasks and datasets in detail. Firstly, we present the statistics of all datasets in Table~\ref{appendix_tab_data}. Then, each task is described as:

\textbf{CoLA} Corpus of Linguistic Acceptability~\cite{warstadt2019neural} is a binary single-sentence classification task to determine whether a given sentence is linguistically ``acceptable''.

\textbf{MRPC} Microsoft Research Paraphrase Corpus~\cite{dolan2005automatically} is a task to predict whether two sentences are semantically equivalent.


\textbf{RTE} Recognizing Textual Entailment~\cite{giampiccolo2007third}, given a premise and a hypothesis, is a task to predict whether the premise entails the hypothesis. 

\textbf{QNLI} Question Natural Language Inference is a binary classification task constructed from SQuAD~\cite{rajpurkar2016squad}, which aims to predict whether a context sentence contains the answer to a question sentence. 

\textbf{CB} CommitmentBank~\cite{de2019commitmentbank} can be framed as three-class textual entailment on a corpus of 1,200 naturally occurring discourses.

\textbf{BoolQ} Boolean Question~\cite{clark2019boolq} is a question answering task where each sample consists of a short passage and a yes/no question about the passage. 

\textbf{WiC} Word-in-Context~\cite{pilehvar2019wic} is a word sense disambiguation task that aims to predict whether the word is used with the same sense in sentence pairs.

\textbf{COPA} Choice of Plausible Alternatives\cite{roemmele2011choice} is a causal reasoning task in which a system is given a premise sentence and must determine either the cause or effect of the premise from two possible choices.

\textbf{SQuAD2.0}  The latest version of
the Stanford Question Answering Dataset~\cite{rajpurkar2018know} is  one of the most widely-used reading comprehension benchmarks that require the systems to acquire knowledge reasoning ability.

\textbf{SWAG} Situations With Adversarial Generations~\cite{zellers-etal-2018-swag} is a task of grounded commonsense inference, which unified natural language inference and commonsense reasoning. It is also widely used to evaluate the ability of PLMs on commonsense knowledge reasoning.

\textbf{Google-RE} The Google-RE corpus contains ~60K facts manually extracted from Wikipedia. The LAMA~\cite{petroni2019language} benchmark manually defines a template for each considered relation, e.g., ``[S] was born in [O]'' for ``place of birth''. Each fact in the Google-RE dataset is, by design, manually aligned to a short piece of Wikipedia text supporting it. There is no training process and during inference, we query the PLMs using a standard cloze template for each relation. It is widely used to probe the model's world knowledge, especially factual knowledge.

\subsection{Hyper-parameters of Fine-tuning}
\label{appendix_parameters}
For fine-tuning, we use the BERT and RoBERTa models as the backbone PLMs and conduct experiments using the open-source toolkit \texttt{fairseq}\footnote{\url{https://github.com/facebookresearch/fairseq}} and \texttt{transformers}\footnote{\url{https://github.com/huggingface/transformers}}. Notably, we apply the same hyper-parameters to all PLMs for simplicity. The training epochs/steps, batch size, and learning rate for each downstream task are listed in Table~\ref{appendix_tab_data}.



\subsection{Does $\mathbb{SE}$ Improve the Robustness?}
\label{appendix_advglue}
Here, we conduct experiments to verify whether $\mathbb{SE}$ improves the robustness of PLMs. In practice, following~\citet{jiang2022rose}, we use the Adversarial GLUE (AdvGLUE)~\cite{wang2021adversarial}, which is a robustness benchmark that was created by applying 14 textual adversarial attack methods to GLUE tasks, to measure the robustness in this study. Table~\ref{tab:advglue} lists the results on all PLMs. With the help of our $\mathbb{SE}$ method, the PLMs achieve consistent improvements on the AdvGLUE benchmark. These results prove that our $\mathbb{SE}$ method is beneficial to the robustness of PLMs.

\input{tables/ablation_threshold.tex}
\subsection{Parameter Analyses on $\mathcal{T}_l$ and $\mathcal{T}_e$}
\label{appendix_parameter_analysis}
As stated in \S\ref{sec:se_method}, we respectively set a threshold $\mathcal{T}_l$ and $\mathcal{T}_e$ for the Correctness-based and Confidence-based metrics to select the hard-to-learn tokens. Here, we analyze the influence of different $\mathcal{T}$ in detail. In practice, taking the $\mathcal{T}_l$ as an example, we train the BERT$\rm_{\texttt{base}}$ with different $\mathcal{T}_l$ (in \{0.05,0.1,0.5,1\}) and evaluate the performance on a combination of GLUE, SuperGLUE (SGLUE for short), SQuAD2.0 and SWAG benchmarks. Table~\ref{tab_ablation_threshold} lists the average scores of these benchmarks. 

Specifically, when the $\mathcal{T}_l$ (i.e., 0.05) is too small, there may be too many easy-to-learn tokens selected by the metric, which could make the PLM pay less attention to the target hard-to-learn tokens and thus slightly affect the efficacy of $\mathbb{SE}$ mechanism. On the other hand, increasing the $\mathcal{T}_l$ makes it hard to learn the few amounts but greatly challenging tokens, thus slightly harming the performance on GLUE/SGLUE. Among them, $\mathcal{T}_l=0.1$ achieves the best, thus leaving as the default setting for correctness-based metric\footnote{We ablate the $\mathcal{T}_e$ spanning \{0.05, 0.1, 0.5, 1, 5, 10\} on the confidence-based metric, and observe the similar trend, where the best setting is $\mathcal{T}_e=1$.}.

\input{tables/statistic_analysis.tex}
\subsection{Analysis of non-complementarity between token-selecting metrics.}
\label{appendix_statistic_analysis}
As aforementioned in the ablation study, costly combining both correctness- and confidence-based metrics to select the tokens in the self-questioning stage does not show complementarity, having not outperformed the default one (correctness-based). 
To explain their non-complementarity, we quantitatively analyze the difference in their vocabulary distributions in Table~\ref{tab:statistic}. 

Specifically, let $P_1$ and $P_2$ denote the token frequency distributions of ``Correctness-based'' and ``Confidence-based'' metrics, respectively. We first use the Jensen-Shannon (JS) divergence~\cite{lin1991divergence} to measure the overall difference between $P_1$ and $P_2$. It can be found that the JS($P_1||P_2$) is only 0.1681, indicating that \textit{\textbf{both distributions are overall similar}}. 
Furthermore, to fine-grained analyze the impact of both distributions on each other, we compute the KL divergence~\cite{kullback1951information} for $P_1\xrightarrow{}P_2$ (i.e., KL($P_2||P_1$)) and $P_2\xrightarrow{}P_1$ (i.e., KL($P_1||P_2$)), respectively. 
Clearly, estimating $P_2$ based on $P_1$ is much easier than the opposite direction, i.e., KL($P_2||P_1$) $<$ KL($P_1||P_2$), indicating that \textit{\textbf{tokens selected by the correctness-based metric contain most of those selected by confidence-based metric}}. These statistics nicely explain the empirical superiority of the correctness-based metric in Table~\ref{tab_ablation_metric}.

\input{tables/all_results.tex}

%% file: tables/dataset.tex
\begin{table*}[h]
\centering
\small
\begin{tabular}{llllcccl}
\toprule
\multicolumn{2}{c}{\textbf{Task}}          & \textbf{\#Train} & \textbf{\#Dev}  & \textbf{\#Class} &\textbf{LR} &\textbf{BSZ} &\textbf{Epochs/Steps} \\ \midrule
\multirow{4}{*}{GLUE}       & CoLA      & 8.5K    & 1,042  & 2   &2e-5  &32  &2668 steps  \\
                            & MRPC        & 3.7K    & 409    & 2   &1e-5  &32  &1148 steps    \\
                            & RTE       & 2.5K    & 278    & 2   &1e-5 &16  &2036 steps     \\ \midrule
\multirow{4}{*}{SuperGLUE}  & BoolQ     & 9.4K    & 3,270  & 2  &1e-5    &16  &10 epochs \\
                            & CB        & 250     & 57     & 2   &2e-5  &16  &20 epochs  \\
                            & WiC        & 6K      & 638    & 2  &2e-5    &16  &10 epochs \\
                            & COPA        &400     &100    & 2  &2e-5  &16  &10 epochs   \\ \midrule
\multirow{2}{*}{Commonsense QA} & SQuAD2.0       &130K    &11,873 & -  &3e-5 &12 & 2 epochs      \\
                            & SWAG &73K    &20K  & - &5e-5 &16 & 3 epochs    \\ \midrule
LAMA & Google-RE & \multicolumn{2}{c}{60K} & - & \multicolumn{3}{c}{N/A} \\
\bottomrule
\end{tabular}
\caption{Data statistics and fine-tuning hyper-parameters of all used tasks in this paper. ``Class'' refers to the label class, ``LR'' means the learning rate and ``BSA'' denotes the batch size. Note that the LAMA benchmark is wrapped into a cloze test to probe the PLM without fine-tuning.}
\label{appendix_tab_data}
\end{table*}

%% file: tables/advglue.tex
\begin{table*}[t]
\centering
\small
{
    \begin{tabular}{lcccccccccccc}
    \toprule
    \multirow{2}{*}{\bf Method} & \multicolumn{6}{c}{BERT$\rm_{\texttt{base}}$} & 
    \multicolumn{6}{c}{\textsc{RoBERTa}$\rm_{\texttt{base}}$} \\
    \cmidrule(lr){2-7} \cmidrule(lr){8-13}
    ~ & RTE &SST-2 & QNLI & MNLI & QQP & \underline{Avg.} & RTE &SST-2 & QNLI & MNLI & QQP & \underline{Avg.} \\
    \midrule
    Baseline & 32.8 & \textbf{29.7} & 40.5 & 22.8 & 38.5 & \underline{32.9} & \textbf{47.1} & \textbf{41.9} & \textbf{36.2} & 21.0 & 30.9 & \underline{35.4} \\
    \textbf{\quad -w/ SE} & \textbf{33.3} & 28.4 & \textbf{42.6} & \textbf{23.5} & \textbf{42.3} & \underline{\textbf{34.0}} & 45.7 & 37.8 & 32.4 & \textbf{23.5} & \textbf{38.5} & \underline{\textbf{35.6}} \\
    \midrule
    \multirow{2}{*}{\bf Method} & \multicolumn{6}{c}{BERT$\rm_{\texttt{large}}$} & 
    \multicolumn{6}{c}{\textsc{RoBERTa}$\rm_{\texttt{large}}$} \\
    \cmidrule(lr){2-7} \cmidrule(lr){8-13}
    ~ & RTE &SST-2 & QNLI & MNLI & QQP & \underline{Avg.} & RTE &SST-2 & QNLI & MNLI & QQP & \underline{Avg.} \\
    \midrule
    Baseline & 45.6 & \textbf{35.8} & 41.4 & \textbf{25.3} & 45.4 & \underline{38.7} & 64.1 & 43.9 & \textbf{61.5} & 33.9 & 44.9 & \underline{49.7} \\
    \textbf{\quad -w/ SE} & \textbf{53.1} & 35.1 & \textbf{45.3} & 24.7 & \textbf{50.0} & \underline{\textbf{41.6}} & \textbf{67.9} & \textbf{48.6} &58.1 & \textbf{34.6} & \textbf{55.1} & \underline{\textbf{52.9}} \\
    \bottomrule
    \end{tabular}
}
\caption{Comparison between $\mathbb{SE}$ and vanilla method applied to all PLMs on AdvGLUE~\cite{wang2021adversarial} benchmark. Average scores on all tasks are \underline{underlined}. The best results are given in \textbf{bold}. 
}
\label{tab:advglue}
\end{table*}

%% file: tables/ablation_threshold.tex

\begin{table}[t]
\centering
\scalebox{0.85}{
\begin{tabular}{lcc}
\toprule
\multirow{2}{*}{Method} & GLUE/SGLUE & SQuAD2.0/SWAG \\ \cmidrule{2-3}
\multicolumn{1}{c}{} & \textit{Avg. ($\Delta$)} & \textit{Avg. ($\Delta$)} \\ \midrule \midrule
BERT$\rm_{\texttt{base}}$ & 74.10 & 74.93 \\ \hdashline
\multicolumn{3}{l}{BERT-SE$\rm_{\texttt{base}}$} \\
\quad $\mathcal{T}_l=0.05$ & 74.63 (\textcolor[RGB]{0,176,80}{+0.53}) & 75.44 (\textcolor[RGB]{0,176,80}{+0.51}) \\
\quad $\mathcal{T}_l=0.1$ & \textbf{75.45 (\textcolor[RGB]{0,176,80}{+1.35})} & \textbf{75.48 (\textcolor[RGB]{0,176,80}{+0.55})} \\
\quad $\mathcal{T}_l=0.5$ & 73.93 (\textcolor{red}{-0.17}) & 75.22 (\textcolor[RGB]{0,176,80}{+0.29}) \\
\quad $\mathcal{T}_l=1$ & 74.02 (\textcolor{red}{-0.08}) & 75.37 (\textcolor[RGB]{0,176,80}{+0.44}) \\
\bottomrule
\end{tabular}
}
\caption{Parameter analysis on the threshold $\mathcal{T}_l$ used in self-questioning stage. The ``Correctness-based'' metric is used in this study. Full results are in Table~\ref{tab:all_results2}.}
\label{tab_ablation_threshold}
\end{table}

%% file: tables/statistic_analysis.tex
\begin{table}[]
    \centering
    \begin{tabular}{ccc}
    \toprule
    \textbf{JS($P_1||P_2$}) & \textbf{KL($P_2||P_1$)} & \textbf{KL($P_1||P_2$)} \\
    \midrule
    0.1681   &0.3875 &0.7506 \\ 
    \bottomrule
    \end{tabular}
    \caption{Distribution difference between vocabulary distributions selected by Correctness-based ``$P_1$'' and Confidence-based ``$P_2$'' metrics. BERT-SE$\rm_{\texttt{large}}$ is used.}
    \label{tab:statistic}
\end{table}

%% file: tables/all_results.tex
\begin{table*}[]
\centering
\small
{
    \begin{tabular}{lcccccccccc}
    \toprule
    \multicolumn{1}{c}{}                         & \bf CoLA                        &\bf  MRPC                                               &\bf  RTE                         & \bf BoolQ         & \bf CB            &\bf  WiC           & \bf COPA          & \multicolumn{2}{c}{\bf Score}      \\
    \cmidrule(lr){2-4} \cmidrule(lr){5-8} \cmidrule(lr){9-10} 
    \multicolumn{1}{l}{\multirow{-2}{*}{\bf Method}} & \textit{Mcc.}               & \textit{Acc.}                             & \textit{Acc.}               & \textit{Acc.} & \textit{Acc.} & \textit{Acc.} & \textit{Acc.} & \textit{\underline{Avg.}} & \textit{$\Delta$} \\
    \midrule \midrule
    \multicolumn{10}{l}{{\textbf{\textit{Baseline PLMs}}}}  \\
    \quad BERT$\rm_{\texttt{base}}$                                & 62.33                        & 88.97                                               & 76.89                        & 75.05          & 85.71          & 66.77          & 63.00          & \underline{74.10}          & -- \\
    \quad BERT$\rm_{\texttt{large}}$                               & 63.00                        & 87.25                                                & 83.80                        & 78.40          & 91.07         & 67.24          & 72.00          & \underline{77.54}          & --              \\
    \hdashline
    \quad RoBERTa$\rm_{\texttt{base}}$                            & 62.00                        & 90.20                                                & 83.12                        & 78.72          & 83.93          & 69.12          & 70.00          & \underline{76.73}          & --              \\
    \quad RoBERTa$\rm_{\texttt{large}}$                            & 64.73                        & 90.69                                                & 88.44                        & 84.37          & 91.07          & 69.90          & 78.00          & \underline{81.03}          & --              \\ \midrule
    \multicolumn{10}{c}{\cellcolor{lightgray}{\textit{``\textbf{Randomly selecting}''}}}  \\
    BERT-SE$\rm_{\texttt{base}}$            &63.23 & 87.01 & 76.17 & 74.83 & 85.70 & 68.00 & 62.00 & \underline{73.85} & \textcolor{red}{-0.25}    \\
    BERT-SE$\rm_{\texttt{large}}$                              & 65.28 & 87.74 & 84.12 & 80.10 & 92.90 & 68.8 & 69.00 & \underline{78.28} & \textcolor[RGB]{0,176,80}{+0.74}  \\
    \hdashline
    RoBERTa-SE$\rm_{\texttt{base}}$                            & 63.78 & 88.73 & 81.59 & 78.83 & 89.29 & 69.43 & 68.00 & \underline{77.09} & \textcolor[RGB]{0,176,80}{+0.36}            \\
    RoBERTa-SE$\rm_{\texttt{large}}$      & 63.42 & 90.20 & 89.89 & 84.13 & 92.86 & 71.00 & 80.00 & \underline{81.64} & \textcolor[RGB]{0,176,80}{+0.61}          \\
    \midrule
    \multicolumn{11}{c}{\cellcolor{lightgray}{\textit{``\textbf{Correctness-based}'' metric}}}  \\
    BERT-SE$\rm_{\texttt{base}}$                         & 63.63 & 89.50 & 77.98 & 74.37 & 89.29 & 67.40 & 66.00 & \underline{75.45} & \textcolor[RGB]{0,176,80}{+1.35}            \\
    BERT-SE$\rm_{\texttt{large}}$                           & 65.66               & 88.23                                               & 85.20               & 80.18          & 92.86          & 68.34          & 78.00          & \underline{79.78}          & \textcolor[RGB]{0,176,80}{+2.24}            \\
    \hdashline
    RoBERTa-SE$\rm_{\texttt{base}}$                         & 62.11                        & 89.71                                             & 84.12               & 79.39          & 92.86          & 71.40          & 74.00          & \underline{79.08}          & \textcolor[RGB]{0,176,80}{+2.35}            \\
    RoBERTa-SE$\rm_{\texttt{large}}$                       & 67.80               & 91.91                                     & 90.25                        & 84.56          & 96.40          & 70.53          & 80.00          & \underline{83.06}          & \textcolor[RGB]{0,176,80}{+2.03}            \\
    \midrule
    \multicolumn{10}{c}{\cellcolor{lightgray}{\textit{``\textbf{Confidence-based}'' metric}}}  \\ 
    BERT-SE$\rm_{\texttt{base}}$                         & 63.17                        & 89.22                                              & 80.51               & 73.98          & 89.29          & 67.24          & 67.00          & \underline{75.77}          & \textcolor[RGB]{0,176,80}{+1.67}            \\

    BERT-SE$\rm_{\texttt{large}}$        & 64.07 & 88.48 & 84.84 & 79.30 & 92.86 & 69.59 & 73.00 & \underline{78.88} & \textcolor[RGB]{0,176,80}{+1.34}            \\
    \hdashline
    RoBERTa-SE$\rm_{\texttt{base}}$                         & 64.06 & 89.71 & 83.03 & 78.10 & 85.71 & 69.44 & 75.00 & \underline{77.86} & \textcolor[RGB]{0,176,80}{+1.13}            \\
    RoBERTa-SE$\rm_{\texttt{large}}$                       & 64.17 & 89.71 & 90.61 & 84.13 & 96.40 & 70.22 & 82.00 & \underline{82.46} & \textcolor[RGB]{0,176,80}{+1.43}           \\
    \bottomrule
\end{tabular}
}
\caption{Full comparison results (corresponding to the average results in Table~\ref{tab_ablation_metric}) between different metrics used to select the hard-to-learn tokens on the combination of GLUE and SuperGLUE benchmarks. ``$\Delta$'' denotes the improvement of $\mathbb{SE}$ methods compared to the baseline PLMs. Average scores on all tasks are \underline{underlined}.}
\label{tab:all_results}
\end{table*}

\begin{table*}[]
\centering
\small
{
    \begin{tabular}{lcccccccccc}
    \toprule
    \multicolumn{1}{c}{}                         & CoLA                        & MRPC                                               & RTE                         & BoolQ         & CB            & WiC           & COPA          & \multicolumn{2}{c}{SQuAD2.0}  &SWAG    \\
    \cmidrule(lr){2-4} \cmidrule(lr){5-8} \cmidrule(lr){9-11} 
    \multicolumn{1}{l}{\multirow{-2}{*}{Method}} & \textit{Mcc.}               & \textit{Acc.}              & \textit{Acc.}               & \textit{Acc.} & \textit{Acc.} & \textit{Acc.} & \textit{Acc.} & \textit{EM} & \textit{F1} &\textit{Acc.} \\
    \midrule 
    RoBERTa$\rm_{\texttt{base}}$                            & 62.00                        & \textbf{90.20}                                                & 83.12                        & 78.72          & 83.93          & 69.12          & 70.00          & 78.79          & 81.92 &79.69              \\
    \hdashline
    \multicolumn{11}{l}{RoBERTa-SE$\rm_{\texttt{base}}$} \\   
    \quad -w/ vanilla LS &63.18 & 89.71 & 83.39 & 78.29 & 89.29 & 69.75 & \textbf{75.00} & 79.01 &82.12 &79.97 \\
    \quad -w/ TLS (Ours) & 62.11                        & 89.71                                         & \textbf{84.12}               & \textbf{79.39}          & \textbf{92.86}          & 71.40          & 74.00          & \textbf{79.41}         & \textbf{82.55} &\textbf{79.88}            \\  
    \toprule
    BERT$\rm_{\texttt{base}}$  &62.33 & 88.97 & 76.89 & \textbf{75.05} & 85.71 & 66.77 & 63.00 & 72.85 & 75.63 & 77.83 \\
    \hdashline
    \multicolumn{11}{l}{BERT-SE$\rm_{\texttt{base}}$} \\
    \quad $\mathcal{T}_l=0.05$ &63.30 & 87.50 & 77.26 & 73.91 & 85.71 & 67.71 & \textbf{67.00} & 72.85 & 75.63 & 77.83 \\
    \quad $\mathcal{T}_l=0.1$ &\textbf{63.63} & \textbf{89.50} & \textbf{77.98} & 74.37 & \textbf{89.29} & 67.40 & 66.00 & \textbf{72.89} & \textbf{75.64} & \textbf{77.91} \\
    \quad $\mathcal{T}_l=0.5$ &61.31 & 88.24 & 77.26 & 73.88 & 85.71 & 67.08 & 64.00 & 72.46 & 75.36 & 77.84 \\
    \quad $\mathcal{T}_l=1$ &62.06 & 88.24 & 77.26 & 73.82 & 82.14 & 68.65 & 66.00 & 72.62 & 75.56 & 77.93 \\
    \bottomrule
\end{tabular}
}
\caption{Full comparison results (corresponding to the average results in Table~\ref{tab_ablation_tls} and~\ref{tab_ablation_threshold}, respectively) on the combination of GLUE, SuperGLUE, SQuAD2.0 and SWAG benchmarks. Best results are given in \textbf{bold}.}
\label{tab:all_results2}
\end{table*}